# Creation and Evaluation of a Pre-tertiary Artificial Intelligence (AI) Curriculum

Thomas K.F. Chiu, Helen Meng, Ching-Sing Chai, Irwin King, Savio Wong, Yeung Yam

*Abstract*—*Contributions:* **The Chinese University of Hong Kong (CUHK)-Jockey Club AI for the Future Project (AI4Future) co-created an AI curriculum for pre-tertiary education and evaluated its efficacy. While AI is conventionally taught in tertiary level education, our co-creation process successfully developed the curriculum that has been used in secondary school teaching in Hong Kong and received positive feedback.**

*Background:* **AI4Future is a cross-sector project that engages five major partners – CUHK's Faculty of Engineering and Faculty of Education, Hong Kong's secondary schools, the government and the AI industry. A team of 14 professors with expertise in engineering and education collaborated with 17 principals and teachers from 6 secondary schools to co-create the curriculum. This team formation bridges the gap between researchers in engineering and education, together with practitioners in education context.**

**Research Questions: What are the main features of the curriculum content developed through the co-creation process? Would the curriculum significantly improve the students' perceived competence in, as well as attitude and motivation towards AI? What are the teachers' perceptions of the co-creation process that aims to accommodate and foster teacher autonomy?**

*Methodology:* **This study adopted a mix of quantitative and qualitative methods and involved 335 student participants.**

*Findings:* **1) two main features of learning resources, 2) the students perceived greater competence, and developed more positive attitude to learn AI, and 3) the co-creation process generated a variety of resources which enhanced the teachers' knowledge in AI, as well as fostered teachers' autonomy in bringing the subject matter into their classrooms.**

*Index Terms*—**Artificial Intelligence (AI) Education, pre-tertiary education, curriculum design, co-creation process, teacher education**

## I. Introduction

THE explosive growth of Artificial Intelligence (AI) is increasingly transforming the way we live, learn and work. To inspire more students to pursue AI in their studies and careers, we need to develop relevant and attractive curriculum at an early age and then following through. [1]. Therefore, teaching AI topics is an important global strategic initiative in educating the next generation [2]. However, AI topics have conventionally been taught in tertiary-level curricula. Naturally, there is a lack of studies about AI curriculum design and development for pre-tertiary education [3].

AI education engages students in exploring how machines can simulate human intelligence, such as to sense, perceive, decide, act, interpret, think, learn and create, as reflected in the proposed curriculum framework in this study. The first idea of teaching children AI was to explore through LOGO programming and the Turtle robot [4], which was designed as learning activities, rather than a curriculum. To date, very few studies have been conducted in formal AI curriculum design for pre-tertiary education [3]. In 2018, A technology company SenseTime published the first textbook series for high schools – *Fundamentals of Artificial Intelligenc*e [5], which focuses on technical content and is designed for students with stronger engineering background. More recently, Massachusetts Institute of Technology [6] examined different hands-on robot learning activities for kindergarten children learning and emphasized the student learning process. There is still the lack of a complete curriculum for AI in pre-tertiary education, because curriculum design necessarily involves the additional elements of pedagogy and assessment.

A "curriculum" refers to all experiences which are planned and guided by teachers, and learned by students, whether it is implemented inside or outside the classroom [7]. It is also a description of what, why, how and when students should learn. How the curriculum is perceived and organized influences the process of teaching and learning [2], [7]. Moreover, teacher autonomy, i.e., the capacity to take control of one's own teaching, is crucial to the teacher's motivation and commitment in providing effective learning opportunities for students on the topic matter [8]. Accordingly, a curriculum that supports teacher autonomy can better optimize learning, especially through the teacher's support of the interests and the needs of the students [8-9].

To actualize a pre-tertiary AI curriculum, a project named AI for the Future (AI4Future) was launched at The Chinese University of Hong Kong (CUHK) to create the first secondary school AI curriculum in Hong Kong. Leveraging the experiences in launching the first AI undergraduate degree program in Hong Kong in 2018, AI4Future is a collaborative project that engages five parties – CUHK's Faculty of Engineering and Faculty of Education, local secondary schools (to which we refer as "pioneering schools"), the local government (HKSARG Education Bureau) and the local AI industry. This project began at the junior secondary level (i.e., Grade 7 – 9) and placed special importance in establishing a collaborative group which involves the team of 14 professors (who are active researchers with expertise covering various branches of AI) together with some 15 postdoctoral fellows, research assistants and undergraduate student helpers, working closely with 17 principals and teachers from 6 pioneering schools in the curriculum's co-creation process. The schools were purposefully selected from different districts of Hong Kong with varied social and economic characteristics; including girls', boys' and coeducational schools with different

.

school banding (that is reflective of overall academic standards) – this aims to address the diverse learning needs of Hong Kong's secondary students. The selected schools have demonstrated that they have placed high emphasis in STEM education and also expressed that they have placed pre-tertiary AI education in high priority for their students.

Our special collaborative team formation is intended to support a special co-creation process for the AI curriculum. The main objective is to generate AI curriculum content that draws on the faculty members' expertise spanning AI's fundamentals, state of the art, to the ethics and societal impact; but adopting a content presentation that the teachers consider palatable and inspirational for junior secondary students. As the co-creation process took form and moved forward, three key observations were noted. First, regular meetings with content presentations and discussions among functional groups and sub-groups began to facilitate teacher professional training for pre-tertiary AI education within the team. Second, iterative refinement of the curriculum content on per-topic and per-module bases involves revisions that span weeks. The revisions tightly integrate efforts from members working across secondary and tertiary education, and aim to provide abundant options for teachers' selection and adaptation as the curriculum enters their classrooms. Such a practice supports teacher autonomy and places the needs of the students at the center of curriculum design. Third, over time, our team forms the foundation of an ecosystem for pre-tertiary AI education. Expansion of this ecosystem is imminent as this project expands with an additional thirty some participating schools in the next phase of the project. The pioneering school teachers have also committed to sharing their experiences and practices in a secondary education context with participating school teachers in the process of expansion.

## II. Drawing reference to K-12 Engineering Education

Since the subject matter of AI is often subsumed under the engineering and computer science curricula in university education, the curriculum in AI4Future draws reference to K-12 Engineering Education when designing a pre-tertiary AI curriculum. Literature suggests that key content for K-12 engineering programs should: 1) be interdisciplinary in its content, 2) include impact and ethical considerations, and 3) foster technical communication skills, engineering thinking, and techniques [10-13]. These are elaborated in the following.

**Interdisciplinary nature**. A well-designed engineering program at the K-12 level should emphasize its interdisciplinary nature to address the absence of interdisciplinary connections in school formal curriculum [11-12]. For example, the program should provide students with opportunities to apply developmentally appropriate mathematics in the context of solving engineering problems. One possible strategy is to allow students to study mathematical concepts through engineering [10].

**Impact and ethics**. The program must expose students to contemporary societal problems that are increasingly complex and interdisciplinary in nature. Students should understand how their proposed solutions to the problems impact life and society locally, globally, economically, environmentally, etc., and also consider the possible ethical issues that could be raised. They have the responsibility to consider the safety and potential effect of the solutions on individual and public [10].

**Technical communication skills**. The program should foster students' communication skills about technical matters. Students should be able to use technical language to explain the processes and output of tools or solutions, and also be able to communicate their technical ideas in everyday language for those without a technological background [10], [13].

**Engineering thinking**. In the K-12 setting, students should be empowered to believe that they can design and troubleshoot solutions to problems and develop new knowledge on their own [10]. They can learn from experience and failure, and suggest improvements over existing solutions. In other words, students should be able to use informed judgment to make decisions about their solutions [12].

**Engineering techniques**. Students should learn and implement various techniques, processes and skills in the program [10], [13]. Techniques refer to step-by-step procedures for specific tasks; processes refer to a series of actions taken to complete a final product; and skills are defined as the ability to perform specific tasks. Therefore, the program should provide students with tools throughout the process of building up their techniques and skills [10].

This project considers the above references informative and some elements are adaptable for our design of a pre-tertiary AI curriculum. Given the boundary-less nature of AI technologies, and pre-tertiary AI education is a brand-new initiative (at least in Hong Kong), our work faces three unique challenges – the first is creation of an AI curriculum that is foundational and specific, to enable a concrete grasp of the topic matter, while opening up broad intellectual horizons for the young (pre-tertiary) students who have yet to decide on their interests and directions for long-term development. Every child is different. Not all are academically gifted; some will do better in one field than in another; but all children should be supported and encouraged to achieve his or her potential. The second challenge lies in translation of this new initiative into practice with available manpower and resources, and there is scarcity of AI talents in all sectors. The third challenge is that the needs in pre-tertiary AI education will vary from one school to another and our work must strive to fulfill all such needs.

To address these challenges, this project aims to design a clear curriculum structure that is modular and reconfigurable, to support flexible learning pathways as needed by various schools. Therefore, an AI curriculum for pre-tertiary education should make space for teachers to recognize each student's personal and cognitive capacities, and to adapt the curriculum to suit the students in their classes [3, 8]. The curriculum should respect differences in ways that different children can best learn, therefore, should provide teachers with the flexibility to ensure that their treatment of the content is appropriate for their student's needs and capabilities. In other words, the curriculum should foster teacher autonomy in designing their own classroom activities / school-based curriculum in leading, assisting and encouraging each student.

Teacher autonomy is an important aspect of the teaching profession [8] that is positively related to perceived self-efficacy and job satisfaction [14-15]. These factors are crucial to teacher motivation, engagement and commitment to fostering effective learning environments for students [16]. This autonomy concerns the relations between the teachers' scope of action and the curriculum's role in providing directions, resources and rules that constrain or extend the learning environment [8].

## III. Curriculum Overview

The project has designed a curriculum framework with 12 chapters and five levels of depth address the pedagogy introduced as illustrated in Table I. The first column lists the various topics, organized as chapters, ranging from the introduction and fundamental concepts (chapters 1-2), various branches of AI e.g., computer vision, speech and language processing (chapters 3-9), as well as societal impact, ethical use and transformation of the future of work (chapters 10-12). These chapters aim to capture breadth and comprehensiveness, and allow teachers to pick and choose appropriate content that fits their teaching objective(s). For example, a teacher may only cover chapters on introduction and society, and perhaps chapters on selected (but not all) branches of AI (e.g., computer vision and machine reasoning, only chapter 3 and 7), yet the curriculum remains coherent and self-contained.

Furthermore, each chapter is organized into five modules: Awareness, Knowledge, Interaction, Empowerment, and Ethics (AKIEE), as elaborated in Table II. These modules can be categorized into the Beginner, Intermediate and Advanced levels, as color-coded in Table II with green, yellow and purple respectively. This module-based, level-up design not only allows flexibility in content selection for the classroom, but also caters for capacity building by offering a clear path to development student AI techniques and skills. In addition, the 5 modules are intended to cover the key elements referred in K-12 Engineering Education mentioned in Section II.

Figure 1 is an infographic that encapsulates the overall curriculum structure. The project began at the core of the illustration, with the introduction of AI and the key drivers of its recent, rapid advancements -- namely, Big Data, Machine Learning and Cloud Computing. Another core emphasis is on ethical considerations in the use of AI applications, as well as related societal impact.

The middle circle in pink illustrates our coverage of various branches in AI, including perceptual machine intelligence such as "see" and "hear", human language technologies such as "speak", "read and write", machine reasoning, use of simulation for problem solving, and how machines can generate content "creatively". The outer circle in green shows various possible applications that are supported by AI, many of which carry important societal implications, especially for the future of work.

## IV. Current Study and Research Design

### A. Research Questions

As mentioned above, this study is the first of its kind in the development of pre-tertiary AI curriculum. Therefore, it poses three research questions that address whether the curriculum can improve the student's perceived competence, attitude and motivation toward AI learning. Accordingly, the research questions (RQ) are:

RQ1: What are the main features of the curriculum content developed through the co-creation process?

RQ2: Would the curriculum significantly improve the student perceived competence, attitude and motivation toward AI learning?

RQ3: What are the teachers' perceptions of the co-creation process that foster teacher autonomy?

TABLE I
CURRICULUM FRAMEWORK

✓ Beginner Unit (BU) ✓ Intermediate Unit (IU) ✓ Advanced Unit (AU)

| Modules \ Teaching Units (✓) | Awareness | Ethics and Impact | Knowledge | Interactions | Empowerment |
|---|---|---|---|---|---|
| 1. Introduction to AI | ✓ | ✓ | ✓ | - | - |
| 2. Fundamentals of AI | ✓ | ✓ | ✓ | ✓ | - |
| 3. "See the World" | ✓ | ✓ | ✓ | ✓ | ✓ |
| 4. "Hear" | ✓ | ✓ | ✓ | ✓ | ✓ |
| 5. "Speak" | ✓ | ✓ | ✓ | ✓ | ✓ |
| 6. "Read" | ✓ | ✓ | ✓ | ✓ | ✓ |
| 7. "AI Reasoning" | ✓ | ✓ | ✓ | ✓ | ✓ |
| 8. "Simulate" | ✓ | ✓ | ✓ | ✓ | ✓ |
| 9. "Think and Create" | ✓ | ✓ | ✓ | ✓ | ✓ |
| 10. Societal Good, Social Impact and Challenges of AI | ✓ | ✓ | ✓ | ✓ | ✓ |
| 11. AI and Ethics | ✓ | ✓ | ✓ | - | ✓ |
| 12. AI and Future of Work | ✓ | ✓ | ✓ | - | ✓ |

TABLE II
ELABORATION ON CURRICULUM MODULES.

| Chapters | Descriptions |
|---|---|
| **Awareness** | Awareness of the history, background and development of various types of AI technologies (corresponding to different subsets of intelligence: machine perception, understanding, reasoning, etc.) |
| **Knowledge** | Identification of key concepts and the impact of AI through eye-catching, illustrative applications, especially usage contexts of local relevance. |
| **Interaction** | Experimentation of AI technologies in AI Lab |
| **Empowerment** | Acquisition of the abilities to design, develop and integrate component AI technologies into end-to-end systems. |
| **Ethics & Impact** | Exploration of AI topics and case studies to promote social good, illustrate transformative effects to the future of work, and reflect on ethical use of AI. |

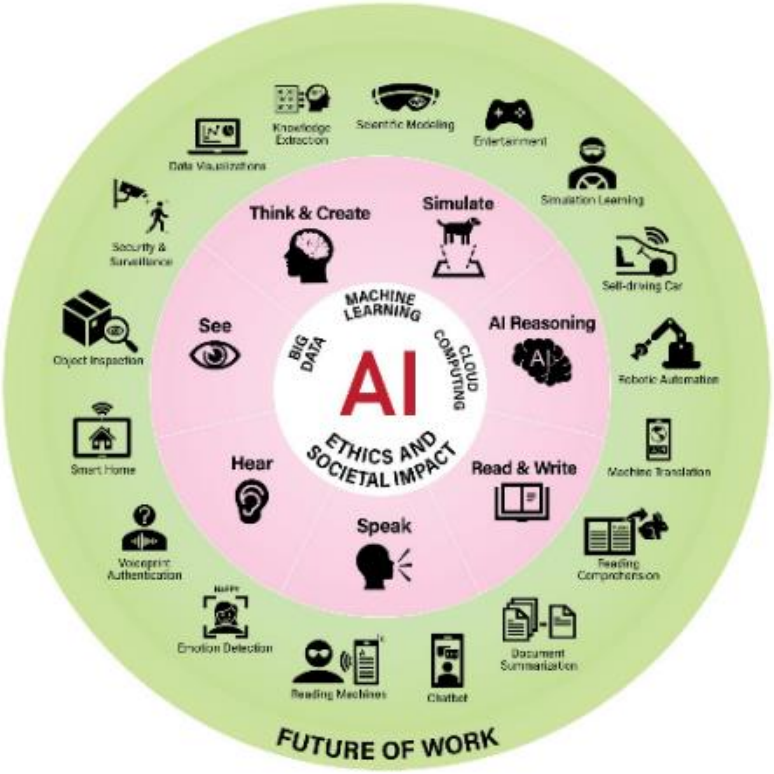

Fig. 1 An infographic providing an overview of the new AI curriculum.

### B. *Research Method*

There are two stages: (1) curriculum development and (2) implementation. In the curriculum development stage, a multilevel co-creation process was adopted – professors from the two faculties authored the technical content based on their research specialization, and then worked with the 17 teachers in AI4future to refine the learning outcomes and pedagogize the content in the year-long regular meetings (biweekly on Monday and Friday afternoons).

In the implementation stage, the teachers considered their school culture, environments and resources, and selected and fine-tuned the relevant content to create learning activities related to the AI technologies, in ways that are most fitting for their students' needs and interests. Due to the COVID-19 pandemic, teachers taught the AI topics in blended environments – both online and face-to-face modes. Moreover, institutional ethical clearance was followed by participant consent. Three hundred and thirty-five students in total, aged 12-16, and 8 teachers from the pioneering schools are involved to evaluate the current the curriculum and its co-creation process. The students completed the online pre- and post-questionnaires; and the teachers participated in individual semi-structured interviews to express how they perceived the co-creation process.

### C. *Measures*

Apart from demographic data, the questionnaire included 5 variables to measure student perceived competence the students' perceived competence, attitude and motivation toward AI. They are *perceived competence* - perceived AI knowledge (AIKG), AI readiness (AIRD), *attitude* - AI confidence (AICF), AI relevance (AIRE) and *motivation* - Intrinsic motivation to learn AI (AIIM). Each variable is measured with a 6-point Likert scale, adapted from previous studies with acceptable reliability and validity [17-19], see Appendix. The items were also checked by three experience educational researchers to make sure that the wording and language were understandable. The following elaborates on the 5 variables:

Perceived AI knowledge (AIKG) – this is newly proposed especially, based on the content design of the project's AI curriculum. This variable measures the student's self-perception of the level of basic knowledge they have acquired for AI. The variable has 4 items, and an example is, "I have general knowledge about how AI are used today.".

AI readiness (AIRD). This variable is adopted from a previous study [17]. It measures the student's perception of the level of comfort in everyday use of various forms of AI technologies. Stronger perception indicates that students hold a favorable viewpoint regarding the adoption of AI in applications. It has 6 items with the original reliability $\alpha$ =.88, and an example is item is "AI technology gives people more control over their own lives."

AI confidence (AICF) [17]. This variable measures the students' perceived confidence in learning the content of AI. The scale has 5 items with the adequate reliability $\alpha$ =.88. An example is "I'm certain that I can succeed if I work hard enough in this AI class."

AI relevance (AIRE). This variable is adopted from a previous study [18]. It measures the students' perception of the relevance of learning AI. It has 6 items with the reliability $\alpha$ = .91, and an example is "The things that I am learning in this AI class will be useful for me.".

Intrinsic motivation to learn AI (AIIM). This variable is adapted from the Motivated Strategies for Learning Questionnaire [19]. The variable has 4 items, with the reliability $\alpha$ =.74, and an example is "In this AI class, I prefer AI topics that arouse my curiosity, even if it is difficult to learn."

In sum, these measures prove to be significant and relevant to measure the learning outcomes as shown in our experimental section

## V. Results and Findings

### A. *Research Question 1 – Main Features*

The first research question targets the analyses of learning resources for each chapter co-created by the professors and teachers that is composed of theory learning, interactive demonstrations, hands-on activities, individual tasks and group discussion, resulting in two main features of the curriculum. The following presents a summary of the main features:

#### 1) *Offer flexible learning pathways*

Flexibility is very important for pre-tertiary education to cater for the diverse needs of the schools and their students. This curriculum aims to offer maximum flexibility for school teachers in adaptation based on their school environments and their students' interests and competencies. Other than the modular and level-up design of the framework, a variety of examples / case studies were also created for the same task or discussion. For example, in the task "Explain why AI technologies may not always work" – three examples, including failure in facial recognition, failure in a chatbot and failure in the prediction of World Cup results are used as illustrations. Moreover, various tools such as Jupyter Notebooks, Blockly, WebAPPs and technologies from industry (e.g., cognitive services, and Google teachable machine) are included for hands-on activities. The project team has also developed a hardware toolkit from scratch – the CUHKiCar (see Fig. 2) is a robotic car with six built-in AI functions, and offers interactive AI experiences to the students. Two AI experiments, namely face-tracking and line following, have been designed and incorporated into the CUHKiCar. They offer introductory experiences in interacting with AI to the students. Furthermore, the CUHKiCar is versatile and can be completely re-programmed with totally new functions in other project work for the students. Overall, this project offers a flexible curriculum with diverse options to match with the students' learning interests and the schools' teaching needs.

#### 2) *Foster local and global understanding*

Student relevance is very important in learning AI. The classroom activities were created such that the tasks enable learning "from local explanations to global understanding". This establishes connections between AI and the students' everyday experiences, i.e., establishes student relevance. In this way, students can gain a better understanding of the societal and personal impacts of AI by combining many high-quality local examples (e.g., KKbox [20], the local subway system's chatbots [21], which are applications of local relevance) that can be extended to understand examples in a global context (e.g., Spotify [22], which is an application found abroad). These examples engage student in a context within which they can develop.

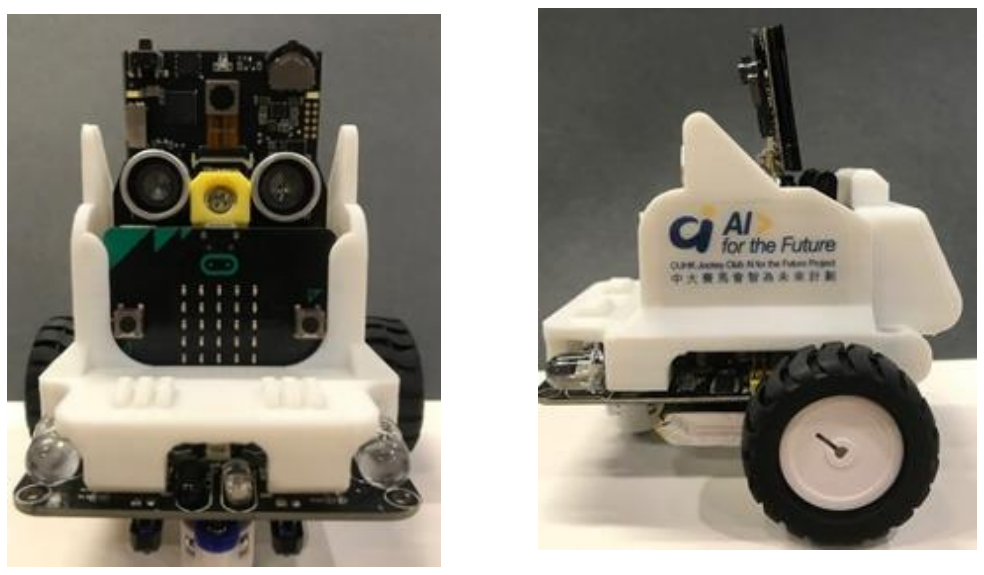

Fig. 2 CUHKiCar.

### *B. Research Question 2 – Student Enhancement*

Paired *t*-tests were adopted to answer RQ2; analyses of covariance (ANCOVA) were used to assess the differences in the post-questionnaire scores after accounting for pre-questionnaire scores [23] to answer RQ2. Table III below presents the descriptive statistics. All the variables with Cronbach alpha coefficients, ranged from 0.88 to 0.92 (> 0.70) were considered internally reliable [24]. Moreover, all the variables met the assumption of homogeneity of variance, with all p values > 0.05 in Levene's tests.

The results of paired *t*-tests were statistically significant, and showed that the students attained higher improvements in all the variables – AIKG, AIRD, AICF, AIRE and AIIM – with t(335) = 8.01 (p<0.001), t(335) =3.45 (p<0.001), t(335) = 4.43 (p<0.001), t(335) = 2.30 (p=0.003) and t(335) = 2.82 (p=0.005), respectively. Therefore, learning with the project's new AI curriculum has effectively enhanced students perceived AI knowledge, AI readiness, AI confidence, AI relevance and intrinsic motivation to learn AI.

TABLE III
DESCRIPTIVE STATISTICS

| **Student (N=358)** | | | | | | |
|---|---|---|---|---|---|---|
| | Pre-questionnaire | | | Post-questionnaire | | |
| Variables | Mean | SD | Cronbach alpha | Mean | SD | Cronbach alpha |
| AIKG | 4.25 | 0.92 | 0.91 | 4.69 | 0.72 | 0.91 |
| AIRD | 4.37 | 0.84 | 0.89 | 4.54 | 0.76 | 0.90 |
| AICF | 4.17 | 0.91 | 0.91 | 4.40 | 0.83 | 0.91 |
| AIRE | 4.52 | 0.86 | 0.90 | 4.67 | 0.71 | 0.88 |
| AIIM | 4.37 | 0.96 | 0.92 | 4.51 | 0.78 | 0.89 |

### *C. Research Question 3 – teacher perception*

This study used teacher training and curriculum features as the framework to analyze 8 teacher interview data to see how the co-creation process foster teacher autonomy, see the following results.

*1)* Teacher training – The analysis revealed that all the participating teachers did not receive formal AI training, and they were able to learn more AI knowledge for curriculum design from the co-creation process. They felt more qualified and confident to teach AI. The following are supporting excerpts from their feedback:

- *'I learned more AI knowledge in the co-design process, which helped me design my AI teaching' (Teacher A)*
- *'In the co-design process, I had chances to try different tools, and discussed with others to learn more AI knowledge. It was fun.' (Teacher M)*

*2) Curriculum features*

The analysis showed that all the teachers agreed that the modular and level-up curriculum design allowed them to adapt the contents to their own effective school-based teaching, through selecting appropriate examples, case studies, tools and modules. This promoted the teachers' sense of autonomy. The following are supporting excerpts from their feedback:

- *'I chose Chapter 1 – awareness, knowledge and ethic and Chapter 7.' (Teacher E)*
- *'I combined the modules from Chapter 1, 11 and 12 to teach my students.' (Teacher C)*
- *'I chose some tools to let student experience what AI is.' (Teacher T)*

## VI. DISCUSSIONS AND CONCLUSION

This article presented three major empirical findings and discussed how this project contributes to pre-tertiary AI education.

### *A. Major Empirical Findings*

The first finding shows that the co-creation process effectively connects the AI and education experts in the university with secondary school teachers. The co-created resources for the curriculum framework result in two main features – offering flexible learning pathways and fostering local and global understanding. Therefore, the process has impact on the development of the resources for classroom practice [25].

The second finding is the proposed curriculum had significant effects on enhancing perceived competence (AIKG, AIRD), for, attitude toward (AICF, AIRE), and intrinsic motivation towards AI (AIIM), see RQ2. This result supports those of related studies that suggests how to design K-12 engineering curriculum, such as those by Delaine et al., [5], Moore et al., [10], and Locke [26], which indicate what key content should be included in effective engineering curricula for schools. This finding further confirms the key content is appropriate for school students, and covers what students should master for AI technologies. The AIKKE curriculum framework is more likely to offer a holistic and comprehensive view of AI, which foster students' knowledge, readiness, confidence, motivation and perceived relevance of AI.

The final finding is that the co-creation process has been shown to be an empowering and enabling process for teachers in supporting their efforts to bring AI into their classrooms. This is accomplished by enhancing the teachers' AI competencies, which in turn, brings out teacher autonomy in shaping the co-created curriculum for their classrooms. The co-creation process not only served as a co-authoring but also offered a contemporary teacher professional development program [27-28].

### *B. Contributions of the AI4Future project*

This study demonstrated that in the AI4Future project, the co-creation process is able to successfully transform the subject matter of AI, which is traditionally taught at the tertiary level, into pre-tertiary, junior secondary classrooms. Figure 4 illustrates the team formation and workflow of the co-creation process. It is worthwhile to note that the co-creation process can: (1) actualize the developing AI curriculum framework by redesigning and pedagogizing content into various classroom learning resources; and (2) enhance the teachers' knowledge AI by offering a sustained professional development process. The various co-created resources also empower the teachers and foster teacher autonomy [29]. Accordingly, this study indicates that this project has guided teachers to inspire students to strive to become future-ready, through facilitating student perceived competence, attitude and motivation towards AI.

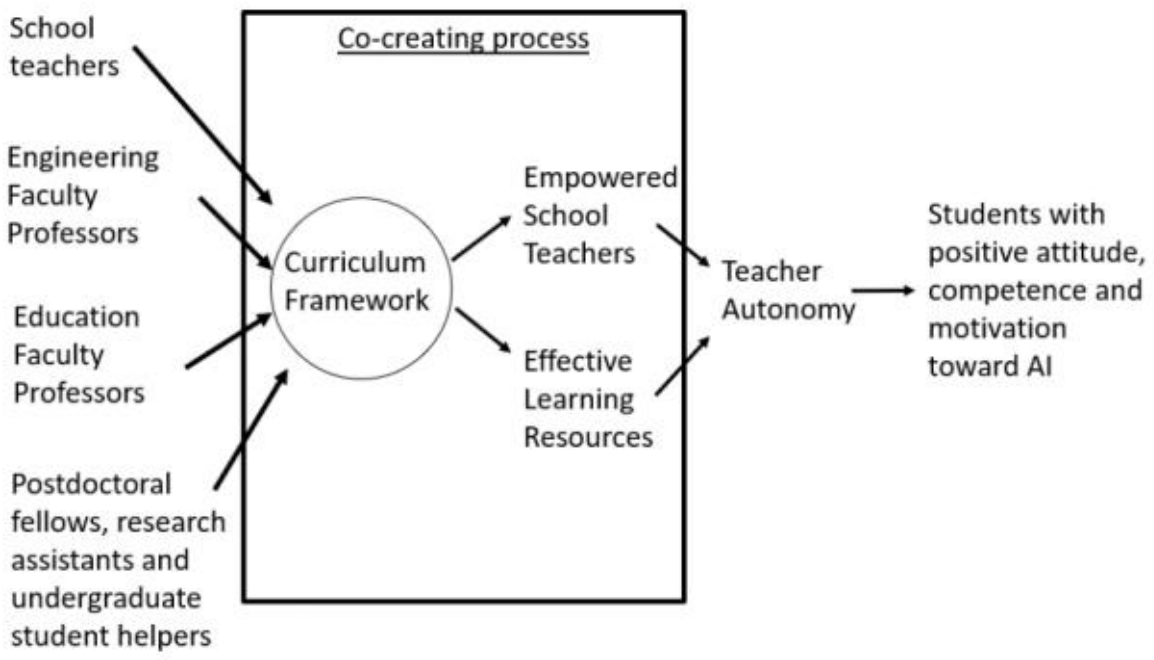

Fig. 4. Team formation and workflow organization of the co-creation process that brings the AI4Future project's new curriculum into junior secondary classrooms.

## VII. Future Opportunities

This paper presents our first step in the creation of a pre-tertiary AI curriculum for Hong Kong, and suggests two main features. The co-creation process enhanced the teachers' knowledge in AI, as well as fostered teachers' autonomy in bringing the subject matter into their classrooms. The students perceived greater competence, and developed more positive attitude to learn AI. Our curriculum has thus far entered its second year of pilot teaching in local pioneering schools. This project is also poised for the next phase and expand to over thirty participating schools. Several valuable opportunities avail to set the stage for our future work. First, as the curriculum enters an increasing number of classrooms, it will be of interest to track with the teachers how content selection and adaptation is done in face of what kinds of students and interests. This will inform further development of learning trajectories for the curriculum. Second, due to the pandemic, pilot teaching as adopted both face-to-face, online and hybrid modes of instruction. Hence, it will be of great interest to see how the curriculum can be optimally adapted to each mode of teaching. Third, the pioneering schools selected for collaboration collectively captures demographic variations, which is well positioned for further research in diversity and inclusion for pre-tertiary AI education [30-31]. Fourth, since AI4Future is a multi-year project, aiming to reach out to the maximum number of secondary schools in Hong Kong, longitudinal research design to track teaching and learning activities will capture valuable data to support learning analytics and inform future pedagogical development.

## Appendix

Pre- and Post- Questionnaire Survey for Students

AIIKG1: I have general knowledge about how AI is used today.
AIIKG2: I have general knowledge about AI capabilities.
AIIKG3: I have general knowledge about AI.
AIIKG4: I have general knowledge of how AI are created. About how AI is created???
AIRD1: AI technologies give people more control over their own lives.
AIRD2: Products and services that use the latest AI technologies are much more convenient to use.
AIRD3: I prefer to use the most advanced AI technologies.
AIRD4: I like AI technologies that allow me to tailor applications to fit my needs.
AIRD5: I find new AI technologies to be mentally stimulating.
AIRD6: I am confident that AI technologies will follow my instructions.
AIRE1: I am aware that AI technology will change the world
AIRE2: The things that I am learning in this AI class will be useful for me.
AIRE3: I should learn the basic knowledge of AI.
AIRE4: It is clear to me how the content of this AI class is related to my future.
AIRE5: The content of this AI class is relevant to my interests.
AIRE6: I could relate the content this AI class to things that I have seen, done or thought in my own life.
AICF1: I feel confident that I will have a good grade in this AI class.]
AICF2: I am certain that I can succeed if I work hard enough in this AI class.
AICF3: I am certain that I can understand the most difficult material presented in this AI class.
AICF4: I am certain that I can learn the basic concepts taught in this AI class.]
AICF5: I am certain that I can understand the most complex material presented by teacher in this AI class.
AIIM1: In this AI class, I prefer AI topics that arouse my curiosity, even if they are difficult to learn.
AIIM2: In this AI class, I prefer the materials that really challenge me so that I can learn new things.
AIIM3: The most satisfying thing for me in this AI class is trying to understand the content as thoroughly as possible.
AIIM4: I like studying in this AI class.

## Acknowledgment

The authors would like to thank the students and the teachers from Diocesan Girls' School, HKSKH Bishop Hall Secondary School, Man Kwan Pak Kau College, Munsang College, The Chinese Foundation Secondary School and Ying Wa College for their involvement in this project AI for future. Special thanks to the principals for their cooperation.

This research is supported (in part) by the Hong Kong Jockey Club Charities Trust.

**Thomas K.F. Chiu** received the BSc degree in Mathematics and Computer Sciences from The Hong Kong University of Science and Technology, and MSc degree in Software Engineering from The Hong Kong Polytechnic University, and PGDE in Technology education and PhD degrees in Information Technology in Education from The University of Hong Kong. He was a software engineer and database trainer in Hong Kong, and the School-University Partnership director in The University of Hong Kong. Currently, He is an assistant professor with Faculty of Education, The Chinese University of Hong Kong, Hong Kong where he is the associate director of Centre for Learning Sciences and Technologies. His research interest includes multimedia learning, technology enhanced learning and STEM education.

**Helen Meng (F’13)** received the S.B., S.M., and Ph.D. degrees in electrical engineering from the Massachusetts Institute of Technology, Cambridge, MA, USA. She is currently a Chair Professor at the Department of Systems Engineering and Engineering Management, The Chinese University of Hong Kong (CUHK). She heads the Curriculum Design Team of the AI4Future Project, which aims to design and develop a pre-tertiary AI curriculum for Hong Kong. Helen's research interests include human–computer interaction via multimodal and multilingual spoken language systems, speech retrieval technologies, and computer-aided pronunciation training. She was awarded the first HKSAR Government Research Grants Council's Theme-based Research Scheme Project in the thematic topic of AI in 2019. She has previously served as the Editor-in-Chief of the IEEE Transactions on Audio, Speech and Language Processing, as well as an Elected Member of the IEEE Signal Processing Society Board of Governors. She is a Fellow of the HKCS, HKIE, ISCA and IEEE.

**Ching-Sing Chai** received his B.A. from the National Taiwan University; PGDE and MA from Nanyang Technological University; and his Ed D from the University of Leicester. He served as a secondary Chinese language teacher and head of department, and as an associate professor in Nanyang Technological University. He is currently a professor in the Chinese University of Hong Kong. He has published more than 100 SSCI papers. His research interests include technological pedagogical content knowledge, language learning, STEM education, and teacher professional development.

**Irwin King** is the Chair and Professor of Computer Science & Engineering at The Chinese University of Hong Kong. His research interests include machine learning, social computing, AI, web intelligence, data mining, and multimedia information processing. In these research areas, he has over 300 technical publications in journals and conferences. He is an Associate Editor of the Journal of Neural Networks (NN) and ACM Transactions on Knowledge Discovery from Data (ACM TKDD). He is President of the International Neural Network Society (INNS), an IEEE Fellow, and an ACM Distinguished Member. He has won the ACM CIKM2019 Test of Time Award and also the ACM SIGIR 2020 Test of Time Award for work done in Social Computing. In early 2010 while on leave with AT&T Labs Research, San Francisco, he taught classes as a Visiting Professor at UC Berkeley. He received his B.Sc. degree in Engineering and Applied Science from California Institute of Technology (Caltech) and his M.Sc. and Ph.D. degree in Computer Science from the University of Southern California (USC). Recently, Prof. King has been an evangelist in the use of education technologies in eLearning for the betterment of teaching and learning through the creation of the Knowledge and Education Exchange Platform (KEEP).

**Savio Wong** is an Associate Professor at the Department of Educational Psychology and Director of the Laboratory for Brain and Education at The Chinese University of Hong Kong (CUHK). He received his Bachelor in Cognitive Science and Ph.D. in Psychology from the University of Hong Kong. He was a postdoctoral fellow at the University of Western Ontario and the University of Southern California. He was appointed as Visiting Associate Professor at the Massachusetts Institute of Technology in 2018. He has been studying the neural mechanism of impulsivity and decision-making in adolescents with functional magnetic resonance imaging (fMRI) techniques and developing psychophysiological markers of emotion with wearable sensors.

**Yeung Yam** is the Principal Investigator of the CUHK Jockey Club Al for the Future Project. He received his B.Sc. degree in Physics from CUHK, and his M.S. and Sc.D. degrees in Aeronautics and Astronautics from MIT. He is the Director of the CUHK Shenzhen Research Institute and a Research Professor at the Department of Mechanical and Automation Engineering. He is also the Associate Master of the Lee Woo Sing College of CUHK. Professor Yam served as the Chairman of the Department of Mechanical and Automation Engineering from 2004 to 2011, and the Interim Dean of the Faculty of Engineering of CUHK in 2018. He is currently serving as the Founding Director of the Intelligent Control Systems Laboratory. His research interests include intelligent systems, human skill acquisition for dexterous manipulations, endoscopic-based surgical robots, computational method for controller design, bio-system modelling and analysis, and technology development for industrial applications. He is a Fellow of the IMechE and a Fellow of HKIE. He has published over 200 technical papers in his fields of interest.